\documentclass[letterpaper, 10 pt, conference]{ieeeconf}  % Comment this line out if you need a4paper

\IEEEoverridecommandlockouts                              % This command is only needed if 
\usepackage[T1]{fontenc}
\usepackage{graphicx}
\usepackage{color}
\usepackage{subfigure}
\usepackage{listings}
\usepackage{algorithm}
\usepackage[noend]{algpseudocode}
\usepackage{booktabs}
\usepackage[table]{xcolor}
\usepackage{array}
\usepackage{amsmath}
\usepackage{amssymb}
\usepackage{siunitx}
\usepackage{hyperref}
\usepackage{cleveref}
\usepackage{xurl}
\usepackage{color}

\lstdefinestyle{style}{
    basicstyle=\ttfamily\footnotesize,
    breakatwhitespace=false,         
    breaklines=true,                 
    captionpos=b,                    
    keepspaces=true,                 
    numbersep=5pt,                  
    showspaces=false,                
    showstringspaces=false,
    showtabs=false,                  
    tabsize=2,
    frame=single
}

\title{\LARGE \bf MARLIN: Multi-Agent Reinforcement Learning Guided by Language-Based Inter-Robot Negotiation}
\author{Toby Godfrey$^{1}$,
William Hunt$^{1}$, and
Mohammad D. Soorati$^{1}$
\thanks{This work was supported by the EPSRC DTP granted through the School of Electronics and Computer Science at the University of Southampton.
$^{1}$Authors are with the Department of Electronics and Computer Science, University of Southampton, Southampton, SO17 1BJ, UK
        {\tt\small \{t.godfrey, w.hunt, m.soorati\}@soton.ac.uk}}}%
\begin{document}

\maketitle
\thispagestyle{empty}
\pagestyle{empty}

\begin{abstract}

Multi-agent reinforcement learning is a key method for training multi-robot systems. Through rewarding or punishing robots over a series of episodes according to their performance, they can be trained and then deployed in the real world. However, poorly trained policies can lead to unsafe behaviour during early training stages. We introduce Multi-Agent Reinforcement Learning guided by language-based Inter-robot Negotiation (MARLIN), a hybrid framework in which large language models provide high-level planning before the reinforcement learning policy has learned effective behaviours. Robots use language models to negotiate actions and generate plans that guide policy learning. The system dynamically switches between reinforcement learning and language-model-based negotiation during training, enabling more effective exploration. MARLIN is evaluated using both simulated and physical robots with local and remote language models. Results show that, compared to standard multi-agent reinforcement learning, the hybrid approach achieves higher performance in early training with limited impact on the final performance. The code is available at \href{https://github.com/SooratiLab/MARLIN}{\texttt{https://github.com/SooratiLab/MARLIN}}.
\end{abstract}
%
%
%
%%%%%%%%%%%%%%%%%%%%%%%%%%%%%%%%%%%%%%%%%%%%%%%%%%%%%%%%%%%%%%%%%%%%%%%%

\section{Introduction}

Multi-robot systems have emerged as a promising paradigm for tackling complex tasks across many domains, requiring advanced decision-making algorithms to coordinate agent behaviour effectively. Multi-Agent Reinforcement Learning (MARL) provides a powerful framework for learning cooperative or competitive policies through interaction with an environment~\cite{albrechtMultiagentReinforcementLearning2024}. Agents learn through exploration by receiving rewards or penalties based on their performance, gradually improving their policies over repeated training episodes. Proximal Policy Optimisation (PPO) is a widely used reinforcement learning method in robotics due to its stable training behaviour, achieved through a clipped surrogate objective that limits the magnitude of policy updates~\cite{schulmanProximalPolicyOptimization2017}. PPO has been extended to multi-agent systems through Multi-Agent PPO (MAPPO), which introduces a centralised critic during training while maintaining decentralised execution~\cite{yuSurprisingEffectivenessPPO2022}. This framework helps address the non-stationarity challenges commonly encountered in MARL environments~\cite{schroederdewittMultiAgentCommonKnowledge2019}. However, training MARL policies for complex real-world scenarios often requires substantial data and time before reliable performance is achieved. Recently, pre-trained large language models (LLMs) have demonstrated strong reasoning and few-shot capabilities across a variety of tasks~\cite{mandiRoCoDialecticMultirobot2024}. Their prior knowledge and reasoning abilities offer a promising opportunity to enhance the decision-making capabilities of multi-robot systems, particularly during early training stages when reinforcement learning policies lack sufficient experience.

In this work, we propose \textbf{M}ulti-\textbf{A}gent \textbf{R}einforcement \textbf{L}earning guided by Language-Based \textbf{I}nter-Agent \textbf{N}egotiation (MARLIN), a hybrid framework in which agents use LLM-mediated dialogue to negotiate natural-language plans that guide MARL training. By dynamically switching between reinforcement learning and language-model-based planning, MARLIN leverages the reasoning capabilities of LLMs to improve performance during the early stages of training while maintaining the final performance of conventional MARL approaches. To reduce training overhead, the system employs off-the-shelf language models without additional fine-tuning. The proposed approach is evaluated against MAPPO and an LLM-only baseline in cooperative navigation tasks where robots must traverse a narrow corridor and swap positions. Experiments are conducted both in simulation and on physical robots using local and remotely hosted language models.

%%%%%%%%%%%%%%%%%%%%%%%%%%%%%%%%%%%%%%%%%%%%%%%%%%%%%%%%%%%%%%%%%%%%%%%%%%%%%%%%

\section{Related Work}

Multi-robot systems have become more common due to their increased robustness and
flexibility compared to single-robot systems~\cite{dorigoSwarmRoboticsPresent2021}.
MARL is widely used to train coordinated behaviour in such systems, with MAPPO
emerging as a particularly effective algorithm owing to its stable, clipped surrogate
objective and centralised training paradigm~\cite{yuSurprisingEffectivenessPPO2022}.
Separately, LLMs have demonstrated strong few-shot and zero-shot reasoning across a
broad range of tasks~\cite{brownLanguageModelsAre2020,dubeyLlama3Herd2024}, and have
increasingly been integrated into robotic systems~\cite{huntSurveyLanguageBasedCommunication2024}.
When extended with action tokens and visual processing capabilities, they enable
end-to-end robot control with improved semantic
generalisation~\cite{brohanRT2VisionLanguageActionModels2023},
and have also demonstrated promise as high-level planners for multi-robot
systems~\cite{huntConversationalLanguageModels2024}.
Recent work has further explored prompt engineering to improve LLM planning
efficiency, with parallel reasoning frameworks
to reduce hallucination~\cite{yaoReActSynergizingReasoning2023}. When multiple LLM instances interact, factual accuracy and reasoning quality can be improved through debate~\cite{duImprovingFactualityReasoning2023}. This has been
explored in robotics by Pfitzer et al., who combined LLM reasoning with neural
networks to reliably issue high-level tasks to robot
swarms~\cite{pfitzerPromptingRobotTeams2025}. Subsequent work has improved the
correctness of such negotiations through premise development and
recovery~\cite{kuMultiAgentLLMDebate2025}. The intersection of LLMs and MARL remains comparatively underexplored. Sun et al.\
surveyed this landscape and identified the training stage as a promising area for
LLM integration~\cite{sunLLMbasedMultiAgentReinforcement2024}, with subsequent work
demonstrating that LLMs can improve sample efficiency and
explainability~\cite{zhangAdvancingSampleEfficiency2024}. The most directly relevant
prior work includes: CRAFT~\cite{choiCRAFTCoachingReinforcement2025}, in which LLMs
decompose high-level tasks and generate MARL reward functions for each sub-task; the
safe MARL framework of Wang et al.~\cite{wangSafeMultiagentReinforcement2024}, which
translates natural language constraints into semantic embeddings to reduce constraint
violations without degrading performance; Ma et al.~\cite{maThinkSpeakDecide2026} combined MARL with LLMs for economic decision-making; and eSpark~\cite{liuKnowingWhatNot2024},
which uses a generative LLM paired with a critic to produce and evaluate exploration
functions that mask actions and improve sample efficiency. However, eSpark is
restricted to homogeneous agents in dense environments, and all three approaches
rely on centralised LLM components.

MARLIN differs from these works by using dyadic inter-agent negotiation to produce
natural-language plans that guide MARL training. Rather than centralising plan
generation or reward shaping, actions emerge from debate between individual
robot-hosted LLM instances, leveraging their prior knowledge to improve
early-training performance. As MARLIN makes no assumptions regarding robot
morphology, reward structure, or the underlying MARL architecture, it is applicable
to a broader class of multi-robot systems than the aforementioned approaches.

\section{Method}

\begin{figure}[tb]
      \centering
      \subfigure[]{\includegraphics[angle=90, width=0.18\linewidth]{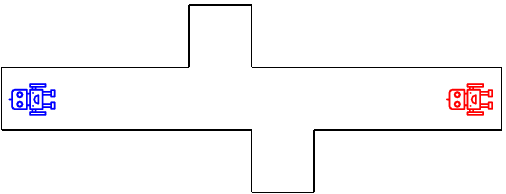}\label{fig-asymmetrical-env}}\hspace{0.25em}
      \subfigure[]{\includegraphics[angle=90, width=0.18\linewidth]{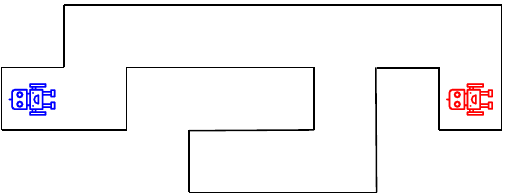}\label{fig-maze-env}}\hspace{0.25em}
      \subfigure[]{\includegraphics[angle=90, width=0.18\linewidth]{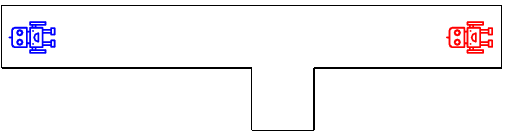}\label{fig-single-env}}\hspace{0.25em}
      \subfigure[]{\includegraphics[angle=90, width=0.18\linewidth]{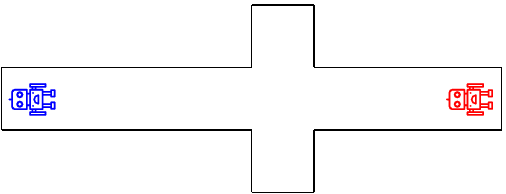}\label{fig-symmetrical-env}}\hspace{0.25em}
      \subfigure[]{\includegraphics[angle=90, width=0.18\linewidth]{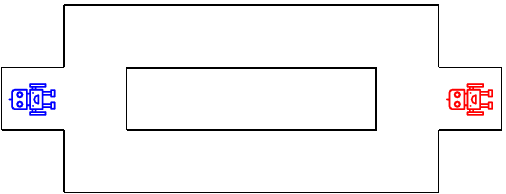}\label{fig-two-env}}
      \caption{Diagrams of the scenarios used for evaluation; Asymmetrical Two Slot Corridor (a), Maze Like Corridor (b), Single Slot Corridor (c), Symmetrical Two Slot Corridor (d), Two Path Corridor (e).}
      \label{fig-env-diagrams}
\end{figure}

% \begin{figure*}[tb]
%     \centering
%     \includegraphics[width=\linewidth]{res/marlin_diagram.pdf}
%     \caption{A diagram of the inter-agent negotiation mechanism. Both agents share the same system prompt, then alternate in suggesting a Top-Level Plan (TLP) and moves for both agents. Critiques can be provided to improve moves until the pair agrees to them. Moves are then simulated, and the next moves are discussed.}
%     \label{fig-negotiation-diagram}
% \end{figure*}

MAPPO is used as the base MARL model for several reasons. It achieves stable training due to the clipping of the surrogate objective function~\cite{yuSurprisingEffectivenessPPO2022}. It also follows the centralised training, decentralised execution paradigm which has been shown to achieve state-of-the-art performance. MAPPO is commonly used in multi-robot systems, allowing MARLIN to be applicable to a wider range of systems. As the agents are cooperative, parameters are shared between agents to decrease the number of trainable parameters. MAPPO is extended with an additional method for action generation to increase performance throughout training. Our system is composed of two action-generation methods: (1) sampling the MARL action distribution (standard MAPPO action selection), and (2) natural language inter-agent negotiation (LLM planner). At each inference step, one of these methods is used, and the generated actions control the robots during training. Situations where LLMs are likely to generate superior plans compared to the action distribution are capitalised on through runtime switching between action-generation functions. Irrespective of how actions are generated, the MARL model benefits by training on the trajectory. Both the standard MAPPO algorithm and our novel approach were implemented using the Ray RLlib library~\cite{therayteamRLlibIndustryGradeReinforcement2024}. In our work, no fine-tuning is performed on the language models. Fine-tuning is likely to yield performance gains, but requires additional time and data, and so is left as a future avenue for exploration.

The system can be modelled as a decentralised partially observable Markov decision process $\langle\mathbf{I}, \mathbf{S}, \{\mathbf{A}_i\}_{i\in\mathbf{I}}, T, \{\Omega_i\}_{i\in\mathbf{I}}, O, \mathcal{R}\rangle$, where $\mathbf{I}=\{1,\dots,n\}$ is the set of agents and $\mathbf{A}_i=\{W,F,B,L,R\}$ denotes the discrete action set corresponding to waiting, forward, backward, left, and right movements, with $\mathbf{A}=\times_{i\in\mathbf{I}}\mathbf{A}_i$ as the joint action space. Each agent observes $o_i=(x,y,x_g,y_g)\in\mathbf{\Omega}_i\subseteq\mathbb{Z}^4$, representing its current position and goal location. The reward function $\mathcal{R}:\mathbf{S}\times\mathbf{A}_i\rightarrow\mathbb{R}$ encourages progress towards the goal while penalising collisions, defined per agent as $\mathcal{R}_i(s_i,a_i)=w\cdot\max(0,1-\frac{d_i}{D_i})-\rho$, where $d_i$ is the Manhattan distance between agent $i$ and its goal, $D_i$ is the initial distance to the goal, $w$ weights progress relative to penalties, and $\rho$ represents collision penalties (fixed to $w=2$ and $\rho=0.5$ for the experiments). The environment consists of narrow corridors in which two agents cannot move past each other, and a way for the agents to either let the other pass or move around each other. Diagrams of the environments can be seen in Fig.~\ref{fig-env-diagrams}. The Symmetrical Two Slot Corridor (see Fig.~\ref{fig-symmetrical-env}) is based on the environment used by Bettini et al.~\cite{bettiniHeterogeneousMultiRobotReinforcement2023}. This idea was expanded to other corridors for the rest of the environments. The goal of all environments is for the agents to swap positions from one end of the corridor to the other. The task was kept simple to aid in reproducibility. A feedforward NN is trained to generate actions for each robot. Each robot has a NN with a $\mathbb{Z}^4$ input vector of its $x$-position, $y$-position, target $x$-position, and target $y$-position; two fully connected hidden layers, each with 256 units and $\tanh$ activation functions, and a $\mathbb{R}^5$ output vector representing the action logits for forwards movement, backwards movement, clockwise turn, counterclockwise turn, and a wait action. MAPPO is used as the underlying MARL algorithm due to its stable training behaviour and effectiveness in cooperative multi-agent tasks~\cite{yuSurprisingEffectivenessPPO2022}. The actor follows the clipped PPO objective, which constrains the deviation between successive policies while incorporating an entropy term to encourage exploration. Training follows the centralised training and decentralised execution paradigm. During execution, each actor operates using only local observations. During training, a centralised critic conditions on global information to estimate the state value. For the two-robot system, the critic receives a $\mathbb{Z}^{13}$ input consisting of each robot’s observation (4-tuple), the other robot’s observation (4-tuple), and the other robot’s one-hot encoded action (5-tuple). The critic is implemented as a feedforward network with a 16-unit $\tanh$ hidden layer and a scalar output representing the value estimate.

\textit{Generation Methods}: A generator function, $G$, is used, such that $G:\mathbf{S}\rightarrow\mathbf{A}^n$ where $\mathbf{S}$ is the set of states, $\mathbf{A}$ is the set of all actions and $n$ is the total number of agents. These generator functions generate the actions for all agents for every step. Our approach uses two distinct generators, Action Distribution Sampling (based on RL), and Inter-Agent Negotiation (using LLMs). \emph {Action Distribution Sampling:} Sampling from the policy's action distribution is the standard way to generate actions for MAPPO. The selected actions depend on the output of the model's network and are based on expected reward, gradually converging towards the best actions for a given state~\cite{suttonReinforcementLearningIntroduction2018}. \emph{Inter-Agent Negotiation:} While generating actions from the model's action distribution works well once the model has adequate experience, it can take significant time and data to reach such a standard. This leaves the model selecting poor-performing actions, especially at the beginning of training or, in the case of a dynamic environment, after the environment changes. By integrating each agent with an LLM instance, they can discuss the next actions to take. The inter-agent negotiation system is extended from previous work~\cite{huntConversationalLanguageModels2024} in which agents take turns discussing and critiquing solutions to a task.

\textit{Inter-Robot Negotiation}:
\begin{algorithm}[t]
    \caption{MARLIN Algorithm}\label{alg-train}
    \begin{algorithmic}[1]
        \Require $\mathbf{N}$ is the set of agents
        \While{$\text{episode} \leq \text{episode}_\text{max}$}
        \State $\overline{p} \gets \text{mean}(\text{pastPerformanceBuffer})$
        \While{$\text{step} < \text{step}_\text{max}$}
        \State $P \gets \text{loadPlan}(\text{state})$
        \State $p_\text{plan} \gets \text{perf}(P)$
        \If{$\text{step} = 0$}
            \If{$\text{episode}<m$}
                \State $G \gets G_\text{ADS}$ \Comment{Initial RL performance}
            \ElsIf{$\text{episode}<2m$}
                \State $G \gets G_\text{IAN}$ \Comment{Initial LLM performance}
            \Else
            \State $G\gets
                        \begin{cases}
                          G_\text{IAN}, & \text{if}\ p_\text{plan} = 1 \\
                          G_\text{IAN}, & \text{if}\ \overline{p} < \overline{p_\text{LLM}} \\
                          G_\text{ADS}, & \text{otherwise} \\
                        \end{cases}$
            \EndIf
        \EndIf
        \If{$P=\varnothing$ \textbf{and} $G=G_\text{IAN}$}
            \State $P \gets \text{makePlan}()$
        \EndIf
        \If{$\text{step} = \frac{\text{step}_\text{max}}{2}$ \textbf{and} $\text{rand}() \leq 0.1$}
            \State $G \gets \text{toggleGenerator}(G)$
        \EndIf
        \State $\text{actions} \gets \text{genActionSingleStep}(G, \text{step})$ %\Comment{Mapping of agents to actions}
        \State $\text{trainStep}(\text{actions})$ \Comment{Update env. \& model}
        \EndWhile
        \State $p \gets \frac{\sum^{\mid\mathbf{N}\mid}_{i=1} \max{(0,1-\frac{d_i}{D_i})}}{\mid\mathbf{N}\mid}$ \Comment{Evaluate performance}
        \State $\text{pastPerformanceBuffer} \gets p$
        \EndWhile
    \end{algorithmic}
\end{algorithm}
To generate actions using language models, robots negotiate the next step through natural-language dialogue, making the planning process interpretable to human observers. The LLMs act as high-level planners rather than controllers; a lower layer translates discrete actions (e.g. \verb|@NORTH|) into linear and angular wheel velocities. As the LLM planners do not attempt heading-dependent movement, their actions are mapped into local actions based on the robot's heading. This abstraction allows MARLIN to be applied to other discrete planning tasks by redefining the available LLM actions and implementing an appropriate action translator. In this work, the LLMs were hosted remotely and accessed over the Internet. At the start of each negotiation round, one robot is designated as the leader---deterministically in our experiments, though random selection is also possible---to reduce the tendency of assistant-style LLMs to defer decisions to other agents~\cite{liCAMELCommunicativeAgents2023}. The robots then alternate messages, proposing and evaluating actions until a joint decision is reached, or a message limit is exceeded. Once agreed, the actions are executed in simulation and the robots receive updated observations. This process repeats until all robots reach their goals or a maximum number of steps is reached. Generated plans are cached for reuse during training unless replaced by higher-performing alternatives. The system prompt specifies the environment rules and required response format, enabling the agreed actions to be parsed automatically. Examples of negotiation are in the GitHub repository. 

\textit{Action Selection Process}: Alg.~\ref{alg-train} summarises the MARLIN training procedure. An initialisation phase of $2m$ episodes establishes baseline performance for the two action generators: Inter-Agent Negotiation ($G_\text{IAN}$) and Action Distribution Sampling ($G_\text{ADS}$). After this phase, the generator used in each episode is selected dynamically based on recent performance and the quality of any cached plan for the current state, thus aiming to select the most performant action generation method. The performance metric compares each agent’s final distance to its goal with its initial distance, such that a successful trial achieves a value of 1. When $G_\text{IAN}$ is used, the system retrieves the best plan for the current state from memory. If no plan exists, or the stored plan performs worse than the recent average, a new plan is generated and cached if it improves performance. To encourage broader exploration during training, there is a 10\% probability that the action generator is switched midway through an episode.

%%%%%%%%%%%%%%%%%%%%%%%%%%%%%%%%%%%%%%%%%%%%%%%%%%%%%%%%%%%%%%%%%%%%%%%%%%%%%%%%

\section{Evaluation}

The effectiveness of our approach is evaluated by testing each component and our combined hybrid system in each scenario. Episode length was limited to 50 moves for all tests, and all common configurations were kept the same between the three systems. The Llama 3.1 8B Instruct model~\cite{dubeyLlama3Herd2024} was chosen as the LLM for all evaluation experiments, and the performance metric from Alg.~\ref{alg-train} was used to evaluate each trial. Models were trained for 1600 episodes. 

\begin{figure*}[t]
  \centering
  \includegraphics[width=\linewidth]{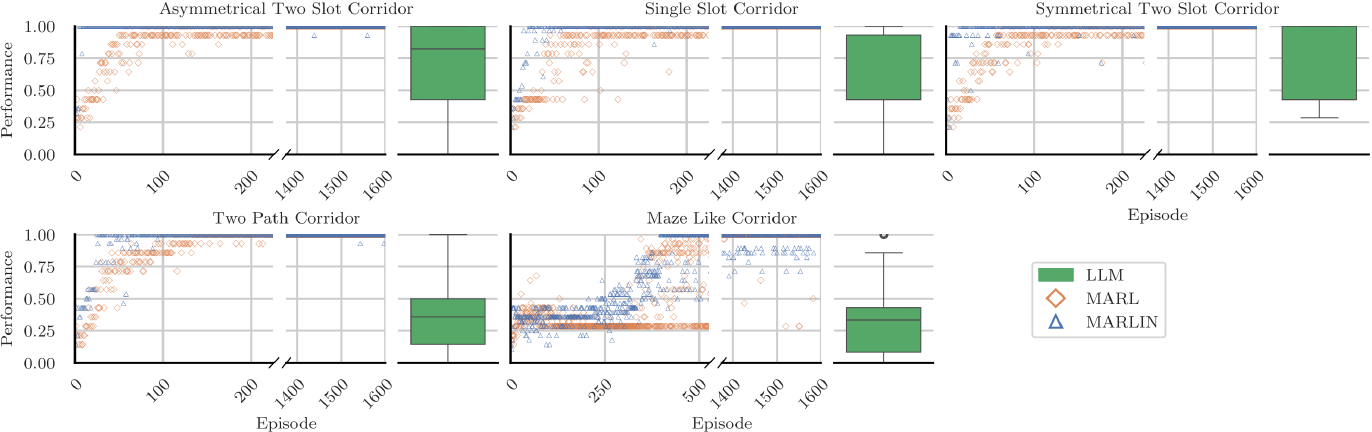}
  \caption{Median performance of the MARLIN and MARL for different scenarios in simulation during the beginning and towards the end of training. The boxplots show the distributions of performance scores for trials using the LLM-based negotiation. %MARLIN reaches highest performance faster than MARL and shows significantly better performance compared to LLM-based negotiation.
  }
  \label{fig-s-perf}
\end{figure*}

\begin{figure}[tb]
    \centering
    \includegraphics[width=0.75\linewidth]{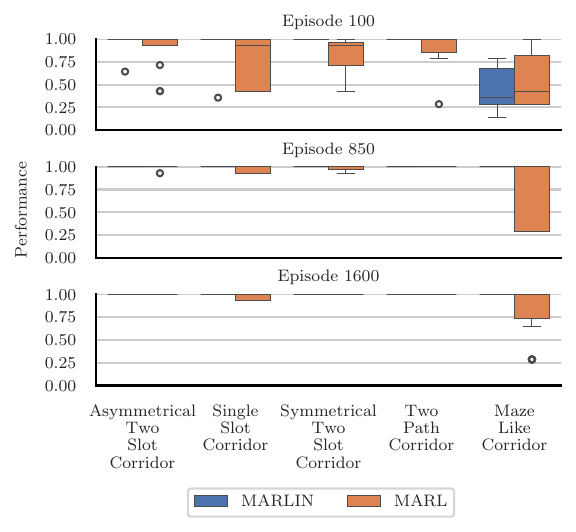}
    \caption{Distribution of performance scores of the MARLIN and standard MARL at different times during training.\label{fig-s-time-dist-perf}}
\end{figure}

\emph{Simulation-Based Evaluation:} The first phase of experimentation was to test the performance of MARLIN against the MAPPO and LLM benchmarks in simulation. Fig.~\ref{fig-s-perf} shows that for all the scenarios, MARLIN reaches highest performance faster than MARL  and performs better compared to the system only using the inter-robot negotiation. In most scenarios, the plans generated by the off-the-shelf LLMs still have difficulty navigating around the other robot. However, robot movement beyond the 50\% performance mark greatly increased the likelihood of task completion. When comparisons are made to the MAPPO benchmark, it is seen that peak performance is reached by MARLIN after fewer training episodes for most scenarios. This shows that plans generated through inter-robot discussion are effective in providing beneficial actions to help train the MARL policy. For the Asymmetrical Two Slot Corridor, Symmetrical Two Slot Corridor, Two Path Corridor, and Single Slot Corridor (see Fig.~\ref{fig-env-diagrams} for the layouts and Fig.~\ref{fig-s-perf} for performance), our method reached peak median performance significantly faster than the MARL. When trying to solve the Maze Like Corridor, both systems achieved similar performance at the start of training, but MARLIN reached peak performance slightly earlier than the MARL system. These data show equal or higher performance by our approach, as compared to a standard MAPPO model, and an increased sample efficiency. This observation highlights how, most of the time, our hybrid approach delivers better performance early in the training phase, than using LLMs alone. Table~\ref{tab-training-perf} shows that at the beginning of training (episode 100) the performance of MARLIN is significantly higher than MARL in all scenarios, apart from the Maze Like Corridor in which MARLIN and MARL performed comparably at the beginning of training. As training progresses, MARL gradually improves and performance approaches that of MARLIN. MARLIN is significantly more performant in the Maze Like and Single Slot corridors, and the other two environments have comparable results. Having statistically significant improvements for the majority of scenarios highlights how MARLIN with LLM-debate can improve in-training performance compared to a typical MARL system. For the Maze Like Corridor MARL and MARLIN had similar performance from the outset. This slower start for MARLIN may be due to the more complex environmental reasoning required by the LLM. In the Two Path Corridor, MARL outperforms MARLIN by the end of training. This is expected and supports our claim, which is to use MARLIN during the early stages of training until MARL has received sufficient experience to perform effectively in the environment. The data in Table~\ref{tab-training-perf} support the trends in Fig.~\ref{fig-s-perf}; initially similar performance between MARL and MARLIN in the Maze Like Corridor, however at episode 850 and later, MARLIN significantly outperformed MARL at the 0.1\% significance level. The latter episodes for the Asymmetrical Two Slot Corridor, MARL and MARLIN have similar performance, indicating that our method does not decrease final performance, however, statistically significant improvements are observed over MARL at the start of training. The same trends of improved initial performance and similar final performance can also be observed in the other scenarios (see Table~\ref{tab-training-perf} and Fig.~\ref{fig-s-perf}). 
\begin{figure}[tb]
    \centering
    \includegraphics[width=\linewidth]{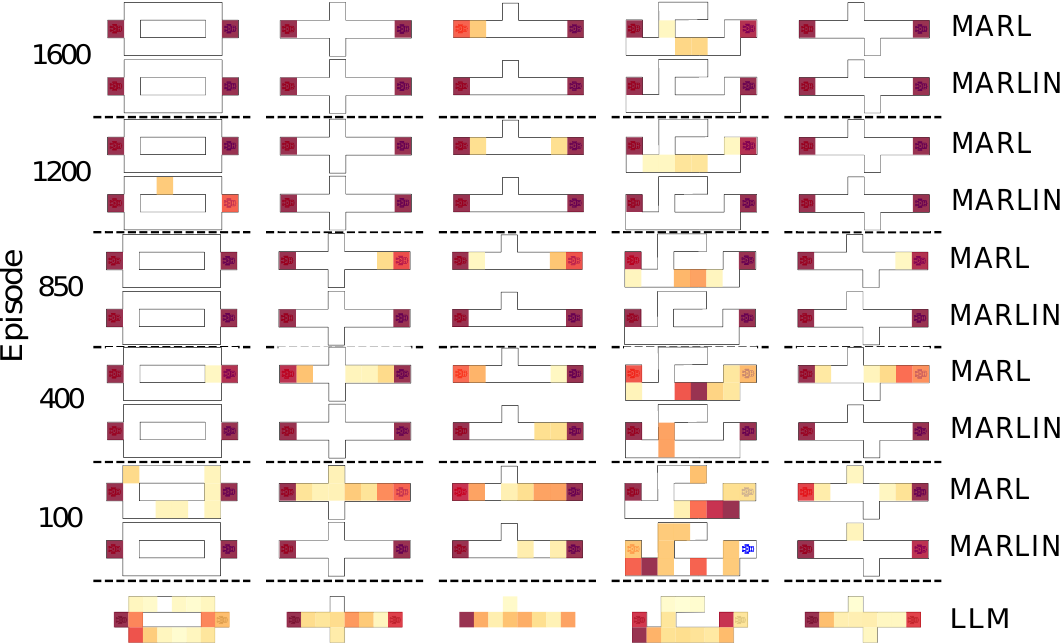}

    \caption{Distribution of end locations throughout training (LLM results are static).    \label{fig-s-heatmaps}}
\end{figure}
As shown in Fig.~\ref{fig-s-time-dist-perf}, initially, at episode 100, our hybrid system has higher median performance than the MARL implementation in almost all the scenarios---the Maze Like Corridor is the only exception, where MARLIN and MARL perform similarly. These data also highlight how the range of performances for our approach is lower than or equal to the MARL benchmark. Towards the end of the training period, MARLIN has successfully solved all five scenarios, compared to MARL which only achieved a perfect median score of 1 in three of the five scenarios. Fig.~\ref{fig-s-heatmaps} also highlights the benefits of MARLIN during training. At the end of earlier episodes, when using MARL agents finish in a wide range of locations for all but the Two Path Corridor, compared to the results when using MARLIN which mostly have a much lower spread of final locations, indicating more trials in which agents reach their goals.

\begin{figure}[t]
    \centering
    \newsavebox{\envfig}
    \savebox{\envfig}{\includegraphics[width=0.4\linewidth]{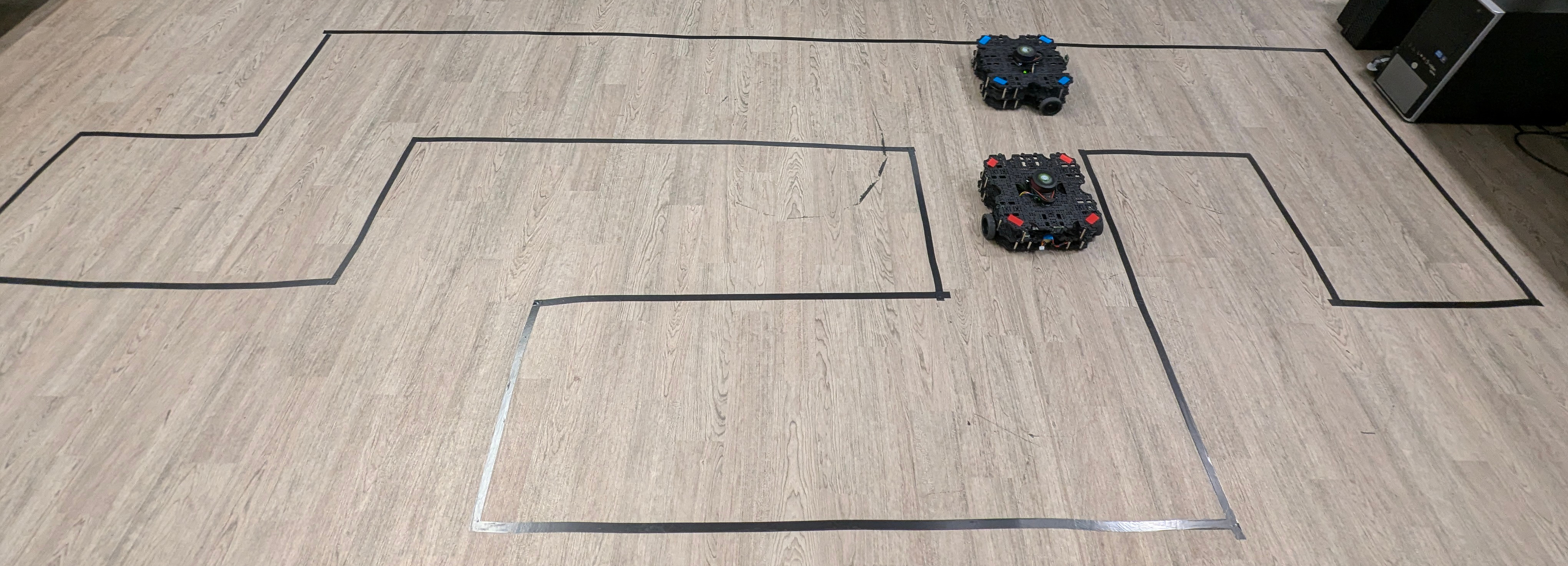}}
    \subfigure[]{\usebox{\envfig}\label{fig-real-env}}\hspace{0.5em}
    \subfigure[]{\includegraphics[height=\ht\envfig]{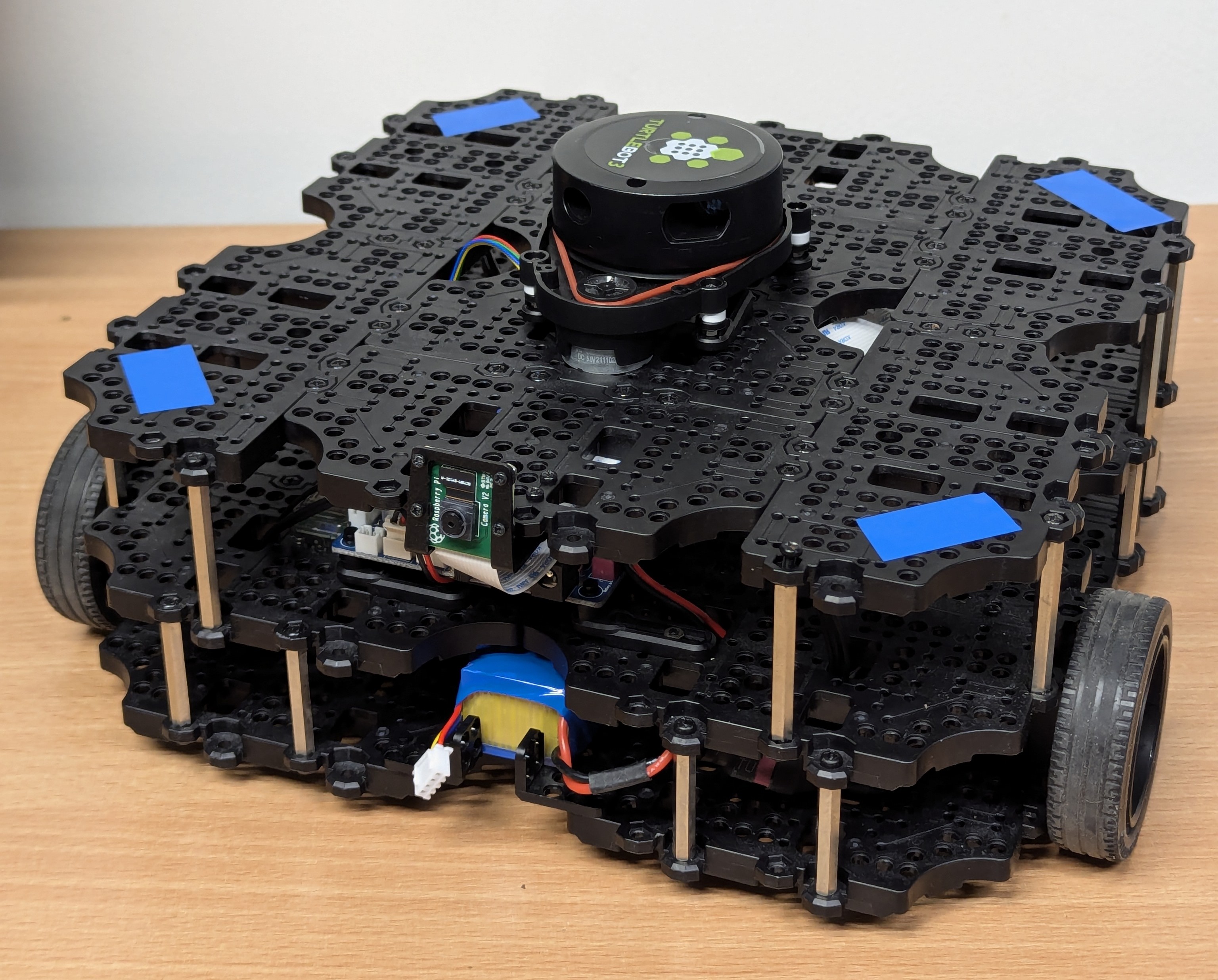}\label{fig-turtlebot}}
    \caption{Environment setup (a) and TurtleBot3 robot platform (b) used for the physical experiments in the Maze Like Corridor. %Results are reported in Fig.~\ref{fig-real-perf}.
    }
    \label{fig-real-env-tb}
\end{figure}

\begin{figure}[tb]
  \centering
  \includegraphics[width=0.9\linewidth]{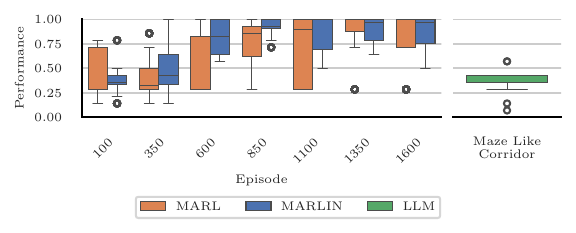}
  \caption{Median performance of the system for the Maze Like Corridor when executed on physical hardware. The box plot shows the performance distribution when the LLM-only system was evaluated on physical hardware.}
  \label{fig-real-perf}
\end{figure}

\emph{Physical Robot Experiments:} To validate the simulation results and reduce the reality gap, experiments were conducted using two TurtleBot3 Waffle robots running ROS2 Humble Hawksbill. The robots use differential drive with a maximum linear speed of \SI{0.26}{\metre\per\second} and rotational speed of \SI{1.82}{\radian\per\second}; position estimation relied on dead reckoning using motor encoder data. The robots communicated with both remote and locally hosted language models via Wi-Fi. Physical testing focused on the Maze Like Corridor scenario (see Fig.~\ref{fig-real-env-tb}), the most challenging environment in simulation. Policy performance for MARLIN and the MARL baseline was evaluated every 250 episodes starting from episode 100, and each trial was repeated 10 times. Results (see Fig.~\ref{fig-real-perf}) show that MARLIN initially improves rapidly and outperforms both the MARL and LLM-only approaches during early training. As in simulation (see Fig.~\ref{fig-s-perf}), MARLIN reaches near-perfect performance earlier, although the MARL baseline eventually matches and slightly exceeds its performance after extended training. These results confirm that LLM-based negotiation can improve early-stage MARL training on physical robots.

\setlength{\tabcolsep}{1.5pt}
\begin{table}[tb]
\centering
\caption{Average performance throughout training. \textbf{Bold} indicates significance ($p<0.001$ over a 50 run paired t-test); %(a) Asymmetrical Two Slot Corridor, (b) Maze-Like Corridor, (c) Single Slot Corridor, (d) Symmetrical Two Slot Corridor, (e) Two Path Corridor.}
\label{tab-training-perf}}
\begin{tabular}{>{\raggedright}p{1.5cm}|cc@{}c|cc@{}c|cc@{}c}
\toprule
\textbf{Scenario} & \multicolumn{3}{c}{\textbf{Ep. 100}} & \multicolumn{3}{c}{\textbf{Ep. 850}} & \multicolumn{3}{c}{\textbf{Ep. 1600}} \\
\cmidrule(r){2-4} \cmidrule(r){5-7} \cmidrule(r){8-10}
& \textbf{MARL} & \textbf{MARLIN} & & \textbf{MARL} & \textbf{MARLIN} & & \textbf{MARL} & \textbf{MARLIN} & \\
\midrule
\rowcolor{gray!15} (a) & 0.8543 & \textbf{1.0000} &  & 0.9814 & 0.9957 &  & 0.9843 & 0.9929 &  \\
(b) & 0.4300 & 0.4271 &  & 0.8529 & \textbf{0.9643} &  & 0.8214 & \textbf{1.0000} &  \\
\rowcolor{gray!15} (c) & 0.7157 & \textbf{0.9071} &  & 0.9329 & \textbf{0.9943} &  & 0.9700 & \textbf{1.0000} &  \\
(d) & 0.8671 & \textbf{0.9714} &  & 0.9929 & \textbf{1.0000} &  & 0.9857 & 0.9952 &  \\
\rowcolor{gray!15} (e) & 0.7771 & \textbf{0.9271} &  & 0.9929 & \textbf{1.0000} &  & \textbf{1.0000} & 0.9667 &  \\
\bottomrule
\end{tabular}
\end{table}

\begin{figure}[tb]
    \centering
    \includegraphics[width=0.5\linewidth]{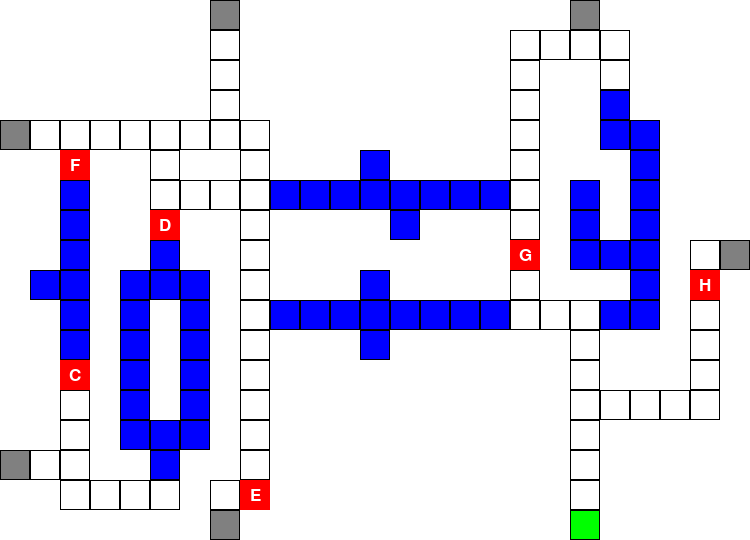}
    \caption{A scenario with a larger number of robots is used to demonstrate our method at scale. Agents negotiate in pairs to pass corridors with no alternative route (e.g. blue areas highlighting our examined corridors). The green cell is the exit, and the grey cells indicate starting positions.}
    \label{fig-large-maze}
\end{figure}

\emph{Scalability Experiments:} To evaluate the scalability of the negotiation system, MARLIN was tested in a warehouse environment shown in Fig.~\ref{fig-large-maze}. Since the negotiation mechanism is designed for pairs of robots, discussions occur only when two agents must coordinate. In this proof-of-concept setup, agents start at grey tiles and move randomly until either reaching the green goal tile and exiting the environment, or encountering another agent in a constrained area where they cannot pass. In such cases, actions are generated through inter-agent negotiation. This experiment demonstrates how dyadic negotiation can be applied to environments with larger numbers of agents by dynamically forming negotiation pairs when required. Pairwise debate has also been used in other multi-agent LLM systems to enable scalability~\cite{kuMultiAgentLLMDebate2025}. Extending this approach to $n$-way negotiation remains an open challenge and is left for future work.

\emph{Local Language Models:} To explore alternative deployment scenarios, local language models were evaluated by allowing agents to negotiate plans for reaching their goals, which were then assessed using the performance metric in Alg.~\ref{alg-train}. The two models tested produced relatively low performance. The \texttt{tinyllama-1.1b-chat} model achieved an average performance of $0.153\pm0.206$, while \texttt{mistral-7b} achieved $0.200\pm0.274$. Due to memory limitations on the TurtleBot3 robots (Raspberry Pi 4 with \SI{4}{\giga\byte} RAM), the models could not run on-board and were instead hosted on a local server queried by the agents. Despite this, performance remained significantly lower than that of the remotely hosted Llama 3.1 8B Instruct model. This is likely due to the much smaller context windows of the local models (2048 tokens for TinyLlama and a 4096-token sliding window for Mistral), which prevents the models from retaining the full negotiation history. Attempts to enforce more structured responses (e.g. JSON or XML) did not significantly improve performance.

%%%%%%%%%%%%%%%%%%%%%%%%%%%%%%%%%%%%%%%%%%%%%%%%%%%%%%%%%%%%%%%%%%%%%%%%%%%%%%%%

\section{Discussion}

This work was developed and evaluated in discrete environments to enable LLMs to reason effectively using grid-based coordinates, as preliminary experiments showed reduced reasoning performance in continuous settings. The Llama 3.1 8B Instruct model was selected due to its availability, performance, and low hosting cost. To improve deployment flexibility, future systems could use distilled models that run directly on robots~\cite{liLargeLanguageModel2025}, provided the performance gap between local and remote models can be reduced. Only cooperative tasks were considered, reflecting the negotiation-based design of MARLIN. Real-world experiments were conducted with two robots due to space and resource constraints. In these tests, the robots functioned largely as physical counterparts of the simulated agents: no onboard sensor data were provided to the system and localisation relied solely on dead reckoning. As a result, encoder noise occasionally caused robots to deviate from the environment, requiring affected trials to be reset and discarded. Finally, negotiation rounds were limited by both message count and token length to maintain concise exchanges.

%%%%%%%%%%%%%%%%%%%%%%%%%%%%%%%%%%%%%%%%%%%%%%%%%%%%%%%%%%%%%%%%%%%%%%%%%%%%%%%%

\section{Conclusions}

MARLIN augments MARL training with inter-agent natural language negotiation. During training, agents use off-the-shelf LLMs to debate and generate actions that guide the MARL policy. Evaluations in simulation and on physical robots show that MARLIN achieves higher performance during early training while often maintaining a comparible final performance as standard MARL. This demonstrates how LLM-mediated dialogue can act as a high-level planner for multi-robot systems, while also improving transparency through natural language reasoning and allowing human preferences to be incorporated via system prompts. Preliminary experiments indicate reduced LLM reasoning performance in continuous environments, suggesting future work should adapt the method for continuous control tasks. Further extensions could explore deploying local on-robot language models while maintaining performance, incorporating additional sensor information and fine-tuning models using expert-generated plans and system parameter tuning. Comparing MARLIN against existing negotiation systems is also an interesting direction for future work.

% \begin{credits}
% \subsubsection{\ackname} This work was done as part of FAST-PI in UKRI Trustworthy Autonomous Systems Hub [EP/V00784X/1] and also supported by UKRI MINDs CDT [EP/S024298/1].

%\subsubsection{\discintname}
%The authors have no competing interests to declare that are
%relevant to the content of this article.
% \end{credits}
\bibliographystyle{splncs04}
\bibliography{references}

\end{document}